%% file: root.tex
\documentclass[a4paper, 10pt, conference]{ieeeconf}      

\IEEEoverridecommandlockouts                              

\usepackage{graphics} 
\usepackage{amsmath} 
\usepackage{amssymb}  
\usepackage{siunitx}
\usepackage[noadjust]{cite}
\usepackage{paralist}
\usepackage{xargs}
\usepackage{tikz}
\usetikzlibrary{angles,patterns,quotes}
\usepackage{tkz-euclide}
\usetkzobj{all}

\usepackage{siunitx}  
\makeatletter
\let\NAT@parse\undefined
\makeatother

\usepackage[nolist,nohyperlinks]{acronym}
\usepackage{algorithm}
\usepackage{algpseudocode}
\usepackage{subcaption}
\usepackage{hyperref}

\title{\LARGE \bf
Anytime Lane-Level Intersection Estimation Based on\\Trajectories of Other Traffic Participants
}

\author{Annika Meyer$^{1}$, Jonas Walter$^{1}$, Martin Lauer$^{2}$ and Christoph Stiller$^{2}$
\thanks{$^{1}$Annika Meyer and Jonas Walter are with FZI Research Center for Information Technology, Karlsruhe, Germany, {\tt\small ameyer@fzi.de}}%
\thanks{$^{2}$Annika Meyer, Martin Lauer and Christoph Stiller are with the Institute of Measurement and Control Systems, Karlsruhe Institute of Technology (KIT), Karlsruhe, Germany}%
}

\newcommand\copyrighttext{%
	\footnotesize \textcopyright 2019 IEEE.  Personal use of this material is permitted.  Permission from IEEE must be obtained for all other uses, in any current or future media, including reprinting/republishing this material for advertising or promotional purposes, creating new collective works, for resale or redistribution to servers or lists, or reuse of any copyrighted component of this work in other works.}
\newcommand\copyrightnotice{%
	\begin{tikzpicture}[remember picture,overlay]
	\node[anchor=north,xshift=10pt,yshift=-10pt] at (current page.north) {\fbox{\parbox{\dimexpr\textwidth-\fboxsep-\fboxrule\relax}{\copyrighttext}}};
	\end{tikzpicture}%
}

\begin{document}
	
\maketitle
\copyrightnotice
\thispagestyle{empty}
\pagestyle{empty}

\begin{acronym}
	\acro{MAP}{maximum a posteriori}
	\acro{MCMC}{Markov chain Monte Carlo}
	\acro{IoU}{Intersection over Union}
\end{acronym}

\begin{abstract}
Estimating and understanding the current scene is an inevitable capability of automated vehicles. Usually, maps are used as prior for interpreting sensor measurements in order to drive safely and comfortably. Only few approaches take into account that maps might be outdated and lead to wrong assumptions on the environment. This work estimates a lane-level intersection topology without any map prior by observing the trajectories of other traffic participants.

We are able to deliver both a coarse lane-level topology as well as the lane course inside and outside of the intersection using Markov chain Monte Carlo sampling. The model is neither limited to a number of lanes or arms nor to the topology of the intersection. 

We present our results on an evaluation set of 1000 simulated intersections and achieve 99.9\% accuracy on the topology estimation that takes only 36ms, when utilizing tracked object detections. The precise lane course on these intersections is estimated with an error of 15cm on average after 140ms. Our approach shows a similar level of precision on 14 real-world intersections with 18cm average deviation on simple intersections and 27cm for more complex scenarios. Here the estimation takes only 113ms in total.

\end{abstract}

\input{content/introduction.tex}
\input{content/point_measurements.tex}
\input{content/tracked_objects.tex}
\input{content/experiments.tex}
\input{content/conlcusion.tex}


\IEEEtriggeratref{19}
\bibliographystyle{ieeetr} 
\bibliography{root}

\end{document}

%% file: content/introduction.tex
\section{Introduction}
In recent autonomous driving systems, highly precise maps have been seen as a inevitable base for not only routing, but also for environment perception \cite{ziegler_making_2014}\cite{Kunz_fusion_2015}. 

Accurate maps allow to replace difficult perception tasks, e.g. recognizing the road boundary, by simple and efficient map lookups.
However, due to construction sites or traffic accidents, the road layout or at least the routing is prone to changes, which leads to outdated maps and a huge effort in updating those maps for ensuring safe and comfortable autonomous driving. In addition, maps rely on a similarly precise localization, which is still an active topic of research. 
Recent system architectures therefore did not only rely on map data, but also added a perception system that is able to estimate further cues on the current road layout \cite{Ballardini_onlineprobabilisticroad_2017,geiger_3d_2014,beck_non-parametric_2014,joshi_generation_2015,topfer_efficient_2015,Meyer_DeepSemanticLane_2018,Dierkes_CorridorSelectionSemantic_2018}.
The majority of these systems focused on lane detection either on simpler (e.g. almost straight) roads or highways \cite{joshi_generation_2015}\cite{topfer_efficient_2015}\cite{Dierkes_CorridorSelectionSemantic_2018}.
They leveraged probabilistic approaches for reasoning over different straight or curved lane hypotheses and relied mainly on visual cues like markings and curb detections as borders. Similarly, in our previous work we applied a deep learning approach to the problem of lane detection in images \cite{Meyer_DeepSemanticLane_2018}. 

For estimating intersections, curbs and markings are not sufficient because the latter usually intersect with each other, leading to highly ambiguous hypotheses. Moreover, both features might be occluded in heavy traffic and approaches that utilize the image coordinate system are often imprecise for distant areas due to perspective distortion.
This makes it especially hard to detect road areas in huge intersections solely based on camera information, e.g. if a lane is represented by just a single pixels.

\begin{figure}[tb]
	\centering
	\includegraphics[width=1\linewidth]{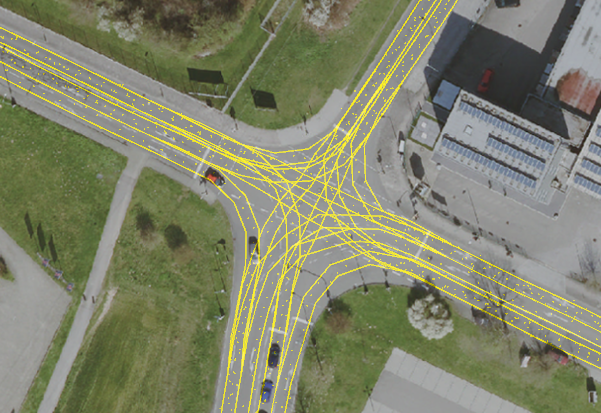}
	\caption{Intersection in Karlsruhe as an example of urban intersections with our estimation. Aerial image source: City of Karlsruhe, www.karlsruhe.de, dl-de/by-2-0}
	\label{fig:example_intersection}
\end{figure}

Hence, the task of estimating intersections has only been researched in a coarse fashion. Currently, autonomous driving systems would fall back to driving with maps assuming their correctness. Previous approaches dealing with the estimation of intersections are presented in the following. 

Beck~\cite{beck_non-parametric_2014} presented an approach that aims at ego lane estimation, but could be extended to intersection estimation. They applied a graph-based shortest path algorithm to find the ego lane in an intersection, that could be applied to all entries of the intersection in order to estimate the lane courses at large. However, they calculated the costs based on marking detections in the image and a coarse semantic labeling, which limits the approach to the image domain and its viewing angle.

Other approaches \cite{Joshi_Jointprobabilisticmodeling_2014}\cite{ruhhammer_crowdsourced_2014} determined, which incoming and outgoing lanes are connected without any further assumption on the geometry of the connection. Joshi and James \cite{Joshi_Jointprobabilisticmodeling_2014} based this kind of estimation on trajectories from other vehicles detected with an onboard Lidar system. Likewise, other approaches \cite{ruhhammer_crowdsourced_2014, chen_probabilistic_2010,Roeth_Roadnetworkreconstruction_2016} used those trajectories for estimating intersections in offline scenarios.
However, those approaches used fleet data and assumed a huge number of trajectory detections per lane.
Ruhhammer et al.~\cite{ruhhammer_crowdsourced_2014} applied clustering algorithms, which themselves require a multitude of trajectory data, whereas Chen and Krumm \cite{chen_probabilistic_2010} calculated a center line for each lane by fitting Gaussian mixture models in order to create a map.
Roeth et al.~\cite{Roeth_Roadnetworkreconstruction_2016} also aimed at map generation, but limited themselves to map graphs instead of lane-level estimation.
Their approach, however, might still be applicable to online lane-level estimation using an extended model. They applied a \ac{MCMC} algorithm, to sample intersection models that are evaluated against the measurement data.

Similarly, Geiger et al. \cite{geiger_3d_2014} estimated the intersection geometry using \ac{MCMC}.
Based on image cues like object detections and tracks, coarse semantic labels and vanishing points. They estimated the intersection structure with a short video sequence approaching the intersection, that took \SI{1}{\second} for estimation. 
The approach achieved promising results although the viewing angle in the dataset was limited to a single camera and lacked sensors facing to the sides. This might have beeen sufficient because of the low complexity of the intersections. 
They published the intersection ground truth for 113 video sequences, but the dataset also has the limitation of a narrow field of view and the lack of lane-level ground truth for intersections with more than a single lane per direction.

Furthermore, the intersection model assumes some limitations that are not consistent with the majority of urban intersections.
At first, they limit the approach to 7 topologies ranging from straight roads to four-armed intersections with only a single lane per driving direction.
Additionally, the crossing arms are forced to be collinear.
Typically, big, urban real-world intersections rather look like that in \autoref{fig:example_intersection}, which is an aerial image with our estimation of an intersection in Karlsruhe.
We analyzed urban roads and highways around Karlsruhe and found that intersections have up to 5 arms and an average of 2.5 arms. Of these, at least \SI{20}{\percent} have more than 3 lanes in total. It should be noted that two structurally separate lanes are considered as two separate arms.

Nevertheless, later works used the dataset of Geiger et al. or their intersection model \cite{Wang_Semanticsegmentationurban_2017}\cite{Ballardini_visuallocalization_2019}, but still could not overcome the drawbacks, that might not hinder simple intersections from being driven autonomously, but still leave more complex ones to be a problem.

Similar to some of the previously mentioned work \cite{Ballardini_onlineprobabilisticroad_2017}\cite{geiger_3d_2014}\cite{joshi_generation_2015}\cite{ruhhammer_crowdsourced_2014}\cite{Roeth_Roadnetworkreconstruction_2016}, we base our approach on the detection and the tracking of other vehicles passing the intersection. Even uncommon intersection structures and spontaneous deviations due to accidents are represented in the measurements.
Additionally, the use of object detections increases the possible viewing range to more than a hundred meters, when working with lidar or radar detections \cite{Winner_AutomotiveRADAR_2016}. 

Like others before \cite{geiger_3d_2014}\cite{Roeth_Roadnetworkreconstruction_2016}\cite{Wang_Semanticsegmentationurban_2017}, we use \ac{MCMC} because it is able to work with contradicting hypotheses in a complex model and can deliver results while approaching an intersection. Using a probabilistic sampling approach like \ac{MCMC}, we are able to incorporate measurements from different sensors with different measurement uncertainties. 
In comparison to all mentioned previous approaches, this work can be applied in an incremental fashion, where the estimation is updated with every measurement and can provide an up-to-date estimate of the intersection at any time in a world-fixed coordinate system (like a map).
Our model also goes a lot further compared to earlier approaches
by not limiting the number of arms or lanes and by providing the exact lane course both within and without the intersection.

In summary, our contributions are the following:

\begin{itemize}
	\item Estimation of intersection topology, geometry and precise lane course by observing other traffic participants
	\item Progressive incorporation of new measurements generating results at any time
	\item Use of an unrestricted model in terms of e.g. number or collinearity of arms 
\end{itemize}

\section{Probabilistic Generative Models}
\label{sec:probs}
For our approach, we rely on the trajectories of other traffic participants as measurements, to estimate the topology and lane course. As depicted in \autoref{fig:steps_diagram}, we are able to use measurements that come from either lidar, camera or radar. For a detailed estimation of the intersection (including the lane course) a tracking system has to be executed beforehand. 

In order to get the best intersection estimation $I$ based on the detections $Z$, we need to calculate the posterior probability $P(I|Z)$.
The model for such an intersection becomes quite complex when regarding each lane and its individual course across the intersection.
Because the posteriors for those models have no simple analytical solution, we use a sampling approach for estimating the intersection.
\ac{MCMC} allows for sampling from such high-dimensional spaces and is especially useful when there is no analytical solution to the problem. 

In the manner of \ac{MCMC} we sample intersections, calculate their posterior probability according to the measurements and decide, whether we would like to retain the solution (or not). During the process we apply simulated annealing \cite{kirkpatrick1983optimization} in order to converge towards the best intersection model. To decide, whether we want retain a new solution, we calculate only the relation between the posterior probabilities of the two intersections $I_a$ and $I_b$ according to the metropolis algorithm \cite{metropolis1953equation}. By applying Bayes' rule, we can simplify our calculations as in (\ref{eq:relation_intersections}) by removing the factor $P(Z)$.
\begin{equation}
\begin{aligned}
\frac{P(I_a|Z)}{P(I_b|Z)} 	
&= \frac{P(Z|I_a) \cdot P(I_a) \cdot P(Z)}{P(Z|I_b) \cdot P(I_b) \cdot P(Z)}  \\
&= \frac{P(Z|I_a) \cdot P(I_a)}{P(Z|I_b) \cdot P(I_b)}
\end{aligned}
\label{eq:relation_intersections}
\end{equation}

Assuming that our measurements $z_i$ are independent of each other, we reformulate (\ref{eq:relation_intersections}) for each intersection $I$ as

\begin{equation}
\begin{aligned}
P(Z|I) \cdot P(I) = P(I)\cdot \prod_{i=1}^{k}P(z_i|I).
\end{aligned}
\label{eq:factor}
\end{equation}

Thus, for our model, we need to calculate the prior $P(I)$ and likelihood $P(z_i|I)$ for every intersection.
When having a deeper look at the measurements in our approach, it's obvious that object detections are in reality not statistically independent. They might belong to the same object and detections belonging to the same object follow the same movements. In order to correctly model the dependency of detections on the same vehicle, we would need to implement an extended object tracking algorithm to calculate the probabilities on the association.
We therefore decided to analyze both options:
\begin{enumerate}[i)]
	\item average points of trajectories (= tracked objects), where the statistical independence can be assumed, and 
	\item single object detections without tracking information, willingly modeling the problem inaccurately. 
\end{enumerate}

%% file: content/point_measurements.tex
\section{Intersection Topology Estimation}
\label{sec:topo}

\begin{figure}[tb]
	\centering	
	\includegraphics[width=\linewidth]{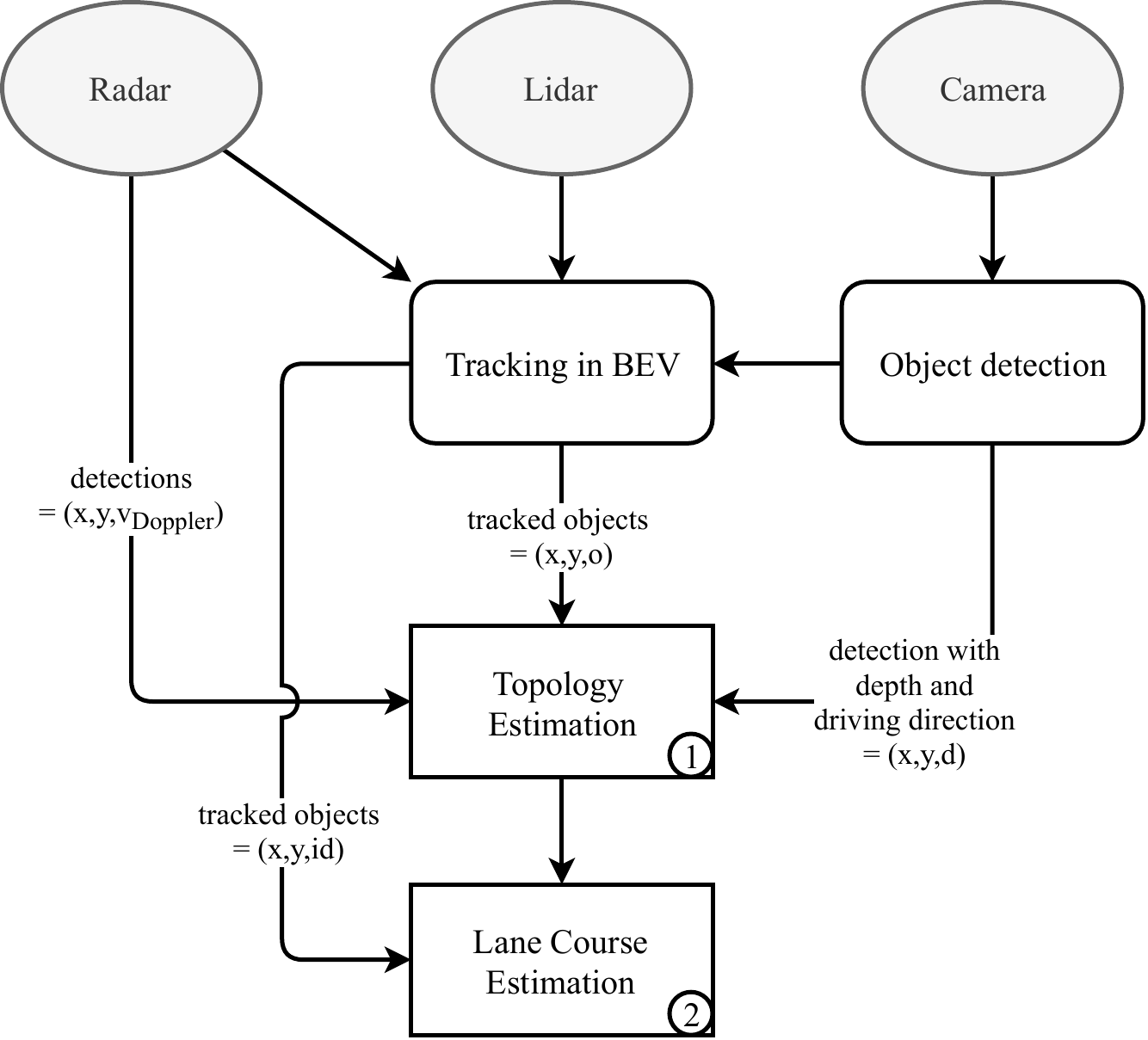}	
	\caption{Overview of the system. Either radar, lidar or camera serve as input. For estimating the lane-level topology \raisebox{.5pt}{\textcircled{\raisebox{-.9pt}{1}}} only point detections and a coarse driving direction are necessary. In order to infer the lane course \raisebox{.5pt}{\textcircled{\raisebox{-.9pt}{2}}}, a tracking system has to be added.}
	\label{fig:steps_diagram}
\end{figure}

The approach is divided into two steps, which makes it possible to process a broad range of sensor measurements as depicted in \autoref{fig:steps_diagram}. 
In the first step, we estimate the lane-level intersection topology. Here, we assume the outgoing lanes to be straight as initial approximation for fast convergence. In the second step, this initial result is refined to the exact course of each lane both inside and outside of the intersection. For the latter, we need all the input data to be processed by a tracking system that calculates the temporal association of the detections. 

\subsection{Preprocessing}
\label{sec:topo_preproc}
The first step requires point detections of other traffic participants (e.g. cars, trucks) that have at least a coarse information about their driving direction (see \autoref{fig:steps_diagram}). 

We are able to use radar measurements, when represented as position $p = (x,y)$ and the Doppler velocity $v_{\text{Doppler}}$ \cite{Winner_AutomotiveRADAR_2016}, since the Doppler velocity of radars can be transformed heuristically into the necessary classification of the driving direction $d$ by 
\begin{equation}
d = 
\begin{cases}
v_{\text{Doppler}} > 0 & \text{leaving}\\
v_{\text{Doppler}} \leq 0 & \text{entering}.
\end{cases}
\end{equation}
This classification assumes, that the perceiving ego vehicle is driving on a lane of the intersection. For detections behind the ego vehicle the classification is vice-versa. 

Using a camera system we would need to preprocess the images with an object detection system, that is able to provide the position of the object in a world-fixed coordinate system and a classification of the driving direction $d$ as entering or leaving the intersection (e.g. \cite{Guindel_Jointobjectdetection_2017}). 

When dealing with lidar detections a tracking system is necessary in order to provide the driving direction of the objects (e.g. \cite{Dewan_Motionbaseddetectiontracking_2016a}). Here, we can reduce the input to detections on the vehicles $p = (x,y)$ and the driving direction as orientation vector $\vec{o}$ derived from the velocity.  
Depending on the data type, a preprocessing for reducing noise and decreasing the number of detections might be useful. E.g. for radar measurements, filtering detections by their compensated Doppler velocity and radar cross-section is necessary in order to avoid static detections and clutter. 

Additionally, a point reduction algorithm like voxelization is required, which combines local neighborhoods into a single detection. Because high resolution radars have a considerable high number of detections on a single vehicle, we can reduce computation times. 

In case of tracked measurements, we speed up the calculations by preprocessing the associated detections. Trajectories passing the intersection are split into two by cutting the trajectory at the closest point to the center. Additionally, we reduce each trajectory to its mean point and mean driving direction, resulting in only one point per trajectory to process. 

\subsection{Intersection Model}
\label{sec:topo_model}

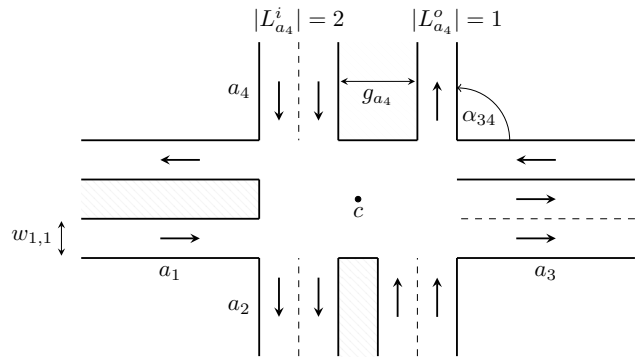
\begin{figure}[tb]
	\centering
	\resizebox{\linewidth}{!}{
		\input{figures/topo_model.tex}
	}
	\caption{The model is fixed at a center point and comprised of a set of arms with assigned lanes. Each lane is assumed to be straight for simplification and is represented by its angle.}
	\label{fig:topo_model}
\end{figure}

The object detections $Z$ are point positions $p = (x, y)$ in a world-fixed coordinate system and their orientation $\vec{o}$ or classification of the driving direction $d$. 

In our approach, we model the intersection $I = (c, A)$ as depicted in \autoref{fig:topo_model}. The model is fixed at a center point $c$
and comprised of a set of arms $a \in A$ with $a = (\alpha_{ab}, g_a, L_a)$. We also model structurally separated lanes, e.g. by a verge or guard rails, with a gap of width $g_a$ between lanes of different driving directions. The angle $\alpha_{ab}$ specifies the angle between adjacent arms $a$ and $b$. Each lane $l \in L_a$ can be classified with $d(l)$ as incoming or leaving lane. Each lane $l$ also has a fixed width $w_l$ assigned.

In our approach, 
we sample new intersections by modifying a single parameter and evaluating, whether this change should be accepted as valid sample or not. 
The detailed process for generating a new sample is enlisted in Algorithm \ref{al:sampling}. We randomly decide which parameter of the intersection is changed using predefined probabilities (sampled by $\omega$). We either modify the number of arms or lanes, the angle of an arm, the center position or the medial strip. When we add an arm, we equally likely add this arm in the largest gap between two others or split one arm into two. 
For the lanes, we add that lane either on the medial strip or at the border of the arm. The same applies for removing a lane. For all add / remove steps, the general geometric bounds of the intersection are regarded (e.g. minimum number of lanes). 

\subsection{Probabilistic Evaluation}
As described in Section \ref{sec:probs}, each sampled intersection $I$ is evaluated based on the conditional probability $P(I|Z)$ given the measurements $Z$. For each detection $z_i$, we need to calculate the likelihood $P(z_i \vert I)$ and for each intersection the prior $P(I)$. 

\subsubsection{Intersection Prior}

The prior probability of the intersection $P(I)$ is based on the number of lanes and the number of arms. An invalid setting of the angles between two arms (e.g. overlapping arms) is already prevented by our sampling procedure (see Algorithm \autoref{al:sampling}).

\begin{algorithm}[tb]
	\begin{algorithmic}[0]
		\State $\omega \leftarrow \mathcal{U}[0,1]$
		\If {$\omega < 0.4$}
		
		{rotate a random arm by $\Delta \alpha \leftarrow \mathcal{U}[\ang{-6},\ang{6}]$}
		\ElsIf {$\omega < 0.6$}	\par	
		shift center by $\{\Delta c,\phi\} \leftarrow \mathcal{U}([\SI{0}{\meter},\SI{6}{\meter}]\times[0,2\pi])$
		\ElsIf {$\omega < 0.7$}\par	
		change gap by $\Delta g \leftarrow \mathcal{U}[\SI{-1.8}{\meter},\SI{1.8}{\meter}]$
		\ElsIf {$\omega < 0.85$}\par
		\State $\theta \leftarrow \mathcal{U}[0,1]$
		\If {$\theta < 0.5$} add arm (details see text)
		\Else \hspace{0.1em} remove arm $a \leftarrow \mathcal{U}_D(A)$
		\EndIf
		\Else 
		\State $\theta \leftarrow \mathcal{U}[0,1]$
		\If {$\theta < 0.5$} add lane (details see text)
		\Else \hspace{0.1em} remove lane $l \leftarrow \mathcal{U}_D(L)$ 
		\EndIf
		\EndIf			
	\end{algorithmic}
	\caption{Sample new intersections by modifying a single parameter at a step.}
	\label{al:sampling}
\end{algorithm}

For the number of arms and lanes, we learned a distribution. The gap $g_a$, the angle between arms $\alpha_{ab}$ and the center $c$ are modeled with non-informative priors. 
Thus, the intersection prior is calculated as 
\begin{equation}
P(I) = P(|A|)  \cdot P(|L|)
\end{equation}

\subsubsection{Likelihoods}
In order to evaluate whether the measurements can be explained given the intersection model, we calculate $P(z_i|I)$. As discussed in Section \ref{sec:topo_preproc}, we have positions of other vehicles $p=(x,y)$. Each position either has a classification of the moving direction $d$ or an orientation $\vec{o}$, which leads to the triplet for each measurement $z_i=(x,y,d)$ or $z_i=(x,y,\vec{o})$. 

Independent of measurement type, we determine the closest lane of the current intersection model with the same driving direction. Using $d_\perp(z_i,M_l)$, we describe the orthogonal distance between the point detection $z_i$ and the center line $M_l$ of that lane. Additionally, we take the angular deviation $d_\sphericalangle(\vec{o}_i,M_l)$ between the center line $M_l$ and the detected orientation $\vec{o}_i$, in case of tracked detections. We assume the following distributions

\begin{equation}
\begin{aligned}
d_\perp(z_i,M_l) & \sim \mathcal{N}(0, \sigma_{\perp})\\
d_\sphericalangle(\vec{o}_i,M_l) & \sim \mathcal{N}(0, \sigma_{\sphericalangle}). \\
\end{aligned}
\end{equation}

We define an association, using the driving direction of both the lane $d(l)$ and the detection $d(z_i)$, as

\begin{equation}
\mathbf{1}(z_i,l) = 
\begin{cases}
0 & d(z_i) = d(l)\\
1 & d(z_i) \neq d(l)
\end{cases}
\end{equation}
determining whether $z_i$ is assigned to lane $l$. When marginalizing over all lanes $L$ in the model, the likelihood $P(z_i|I)$ becomes 
\begin{equation}
P(z_i|I)= \sum_l^L \mathbf{1}(z_i,l) \cdot P_\perp(z_i|M_l,I) \cdot P_\sphericalangle(\vec{o}_i|M_l,I).
\label{eq:topo_z_prob}
\end{equation}

For measurements without an orientation $\vec{o}_i$, we set $P_\sphericalangle(\vec{o}_i|l,I) = 1$.

%% file: figures/topo_model.tex
  \begin{tikzpicture}[
      scale=0.6,
      arrow/.style={thin,<->,shorten >=1pt,shorten <=1pt,>=stealth},
      larrow/.style={thick,<-,>=stealth},
      rarrow/.style={thick,->,>=stealth}
    ]
\draw [thick] (-7,-1.5) -- (-2.5,-1.5)node[midway,below]{$a_1$};
\draw [thick] (-7,1.5) -- (-2.5,1.5);
\draw [thick] (-7,0.5) -- (-2.5,0.5);
\draw [thick] (-7,-0.5) -- (-2.5,-0.5);
\draw [thick] (-2.5,0.5) -- (-2.5,-0.5);

\draw [thick] (7,-1.5) -- (2.5,-1.5)node[midway,below]{$a_3$};
\draw [thick] (7,1.5) -- (2.5,1.5) node (origo) {};
\draw [thick] (7,0.5) -- (2.5,0.5);
\draw [thin, dashed] (7,-0.5) -- (2.5,-0.5);

\draw [thick] (-2.5,-4) -- (-2.5,-1.5)node[align=center,midway,left]{$a_2$};
\draw [thin,dashed] (-1.5,-4) -- (-1.5,-1.5);
\draw [thick] (0.5,-4) -- (0.5,-1.5);
\draw [thick] (-0.5,-1.5) -- (0.5,-1.5);
\draw [thick] (-0.5,-4) -- (-0.5,-1.5);
\draw [thin, dashed] (1.5,-4) -- (1.5,-1.5);
\draw [thick] (2.5,-4) -- (2.5,-1.5);

\draw [thick] (-2.5,4) -- (-2.5,1.5)node[midway,left]{$a_4$};
\draw [thin,dashed] (-1.5,4) -- (-1.5,1.5)node[pos=0,above]{$|L^i_{a_4}|=2$};
\draw [thick] (-0.5,1.5) -- (1.5,1.5);
\draw [thick] (-0.5,4) -- (-0.5,1.5);
\draw [thick] (1.5,4) -- (1.5,1.5);
\draw [thick] (2.5,4) -- (2.5,1.5)node (beta) [pos=0,above]{$|L^o_{a_4}|=1$} ;

\tkzDefPoint(5,1.5){A}
\tkzDefPoint(2.5,1.5){B}
\tkzDefPoint(2.5,2.5){C}
\pic [draw, ->, angle radius=8mm, angle eccentricity=0.6, "$\alpha_{34}$"] {angle = A--B--C};

\draw[opacity=0.2, draw=none, pattern=north west lines, pattern color=black] (-7,-0.5) rectangle (-2.5,0.5);
\draw[opacity=0.2, draw=none, pattern=north west lines, pattern color=black] (-0.5, -4) rectangle (0.5, -1.5);
\draw[opacity=0.1, draw=none, pattern=north west lines, pattern color=black] (-0.5, 4) rectangle (1.5, 1.5);

\draw [arrow] (-0.5,3) -- (1.5,3)node[midway,below]{$g_{a_4}$};
\draw [arrow] (-7.5,-0.5) -- (-7.5,-1.5)node[align=center,midway,left]{$w_{1,1}$};

\draw [rarrow] (-5,-1) -- (-4,-1);\draw [larrow] (-5,1) -- (-4,1);
\draw [larrow] (5,-1) -- (4,-1);\draw [rarrow] (5,1) -- (4,1);
\draw [larrow] (5,0) -- (4,0);

\draw [larrow] (2,3) -- (2,2);
\draw [rarrow] (-1,3) -- (-1,2);\draw [rarrow] (-2,3) -- (-2,2);
\draw [rarrow] (1,-3) -- (1,-2);\draw [rarrow] (2,-3) -- (2,-2);
\draw [larrow] (-1,-3) -- (-1,-2);\draw [larrow] (-2,-3) -- (-2,-2);
\filldraw (0,0) circle (2pt) node[align=right,below] {$c$};

    \end{tikzpicture}

%% file: content/tracked_objects.tex
\section{Lane Course Estimation}

\begin{figure}[b]
	\begin{center}
	\begin{subfigure}{0.492\linewidth}
		\centering
		\includegraphics[width=1\linewidth]{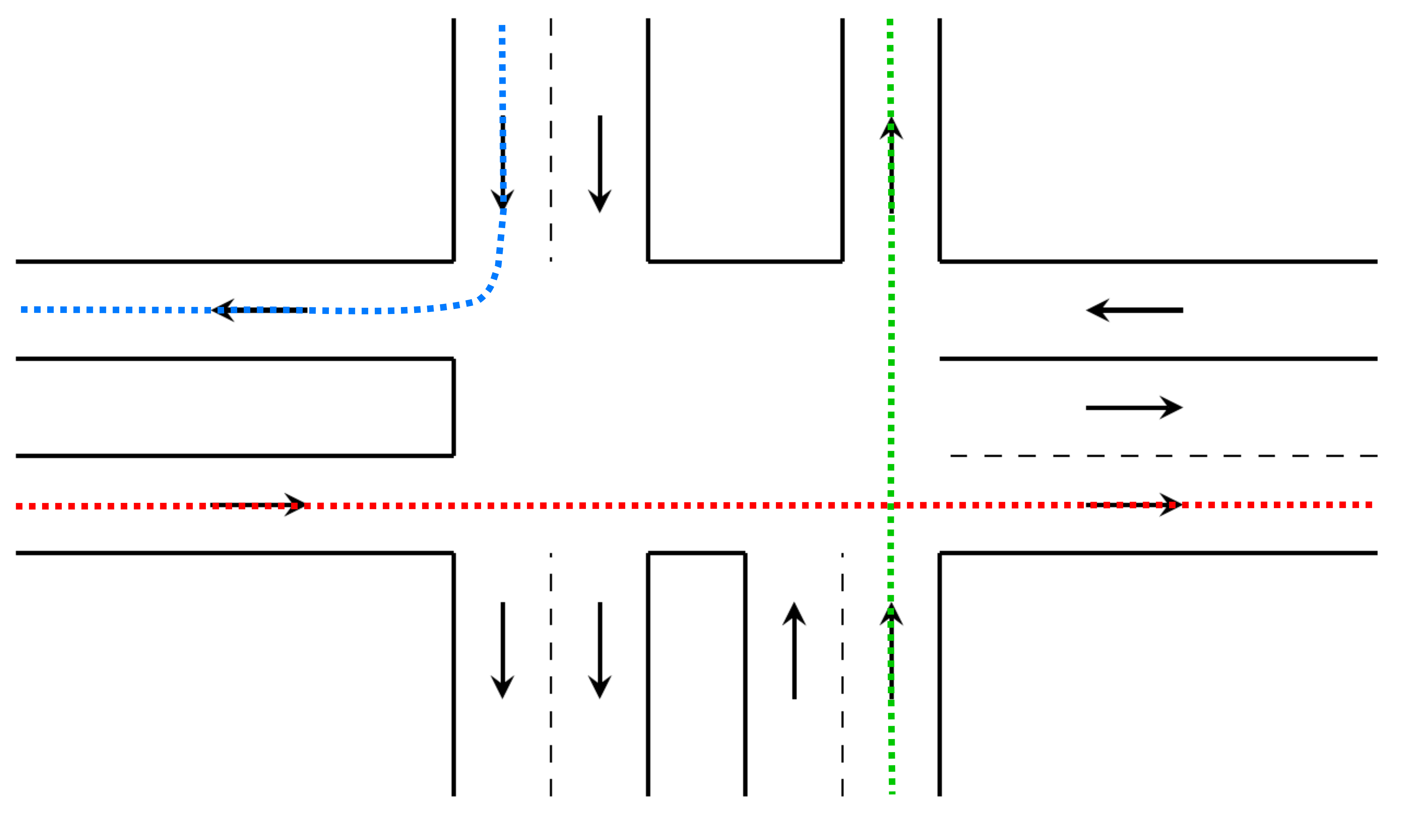}
		\caption{}
		\label{fig:1laneinittraj}
	\end{subfigure}
	\begin{subfigure}{0.492\linewidth}
		\centering
		\includegraphics[width=1\linewidth]{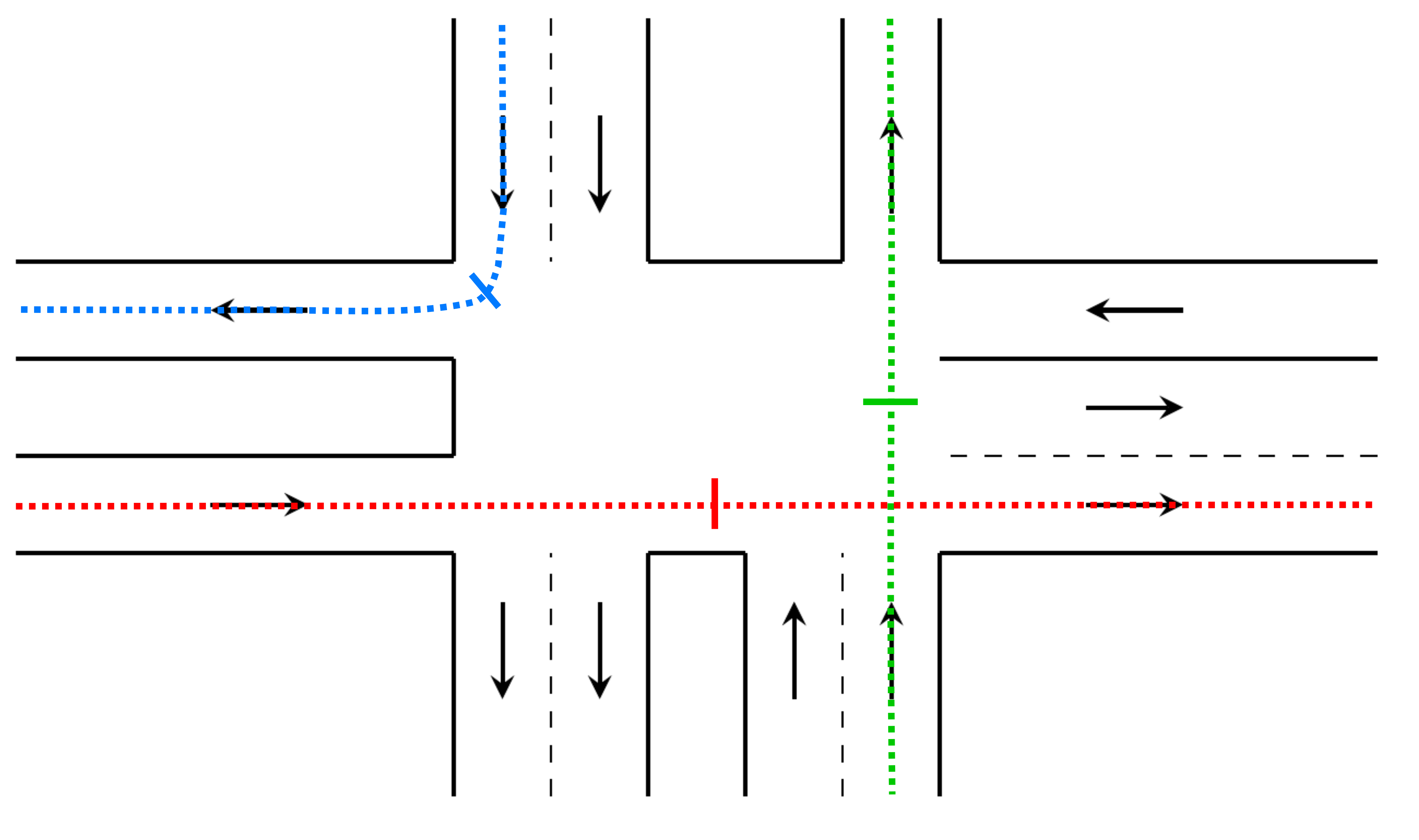}
		\caption{}
		\label{fig:2laneinittrajsplit}
	\end{subfigure}
	\begin{subfigure}{0.492\linewidth}
		\centering
		\includegraphics[width=1\linewidth]{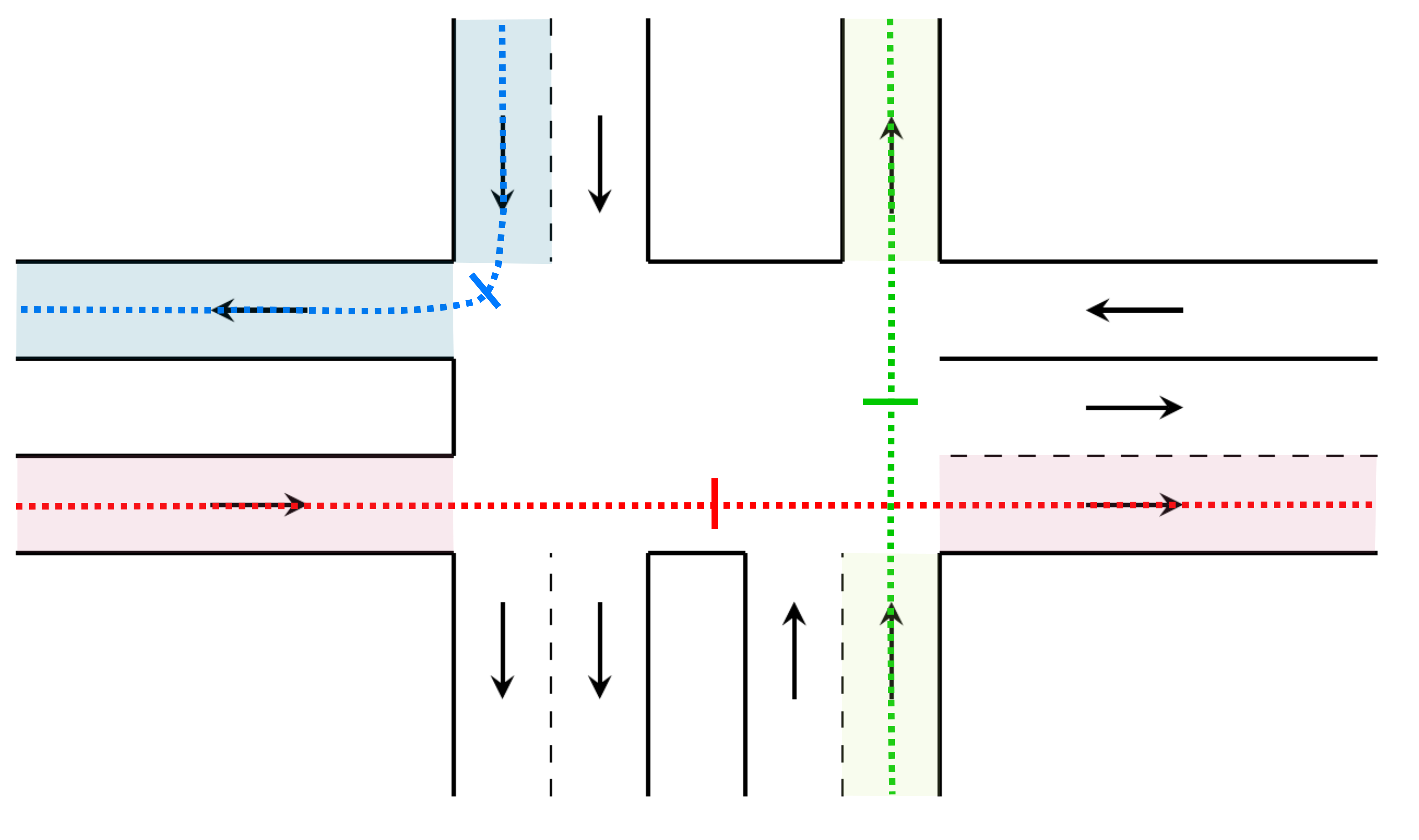}
		\caption{}
		\label{fig:3laneinitsplitareas}
	\end{subfigure}
	\begin{subfigure}{0.492\linewidth}
		\centering
		\includegraphics[width=1\linewidth]{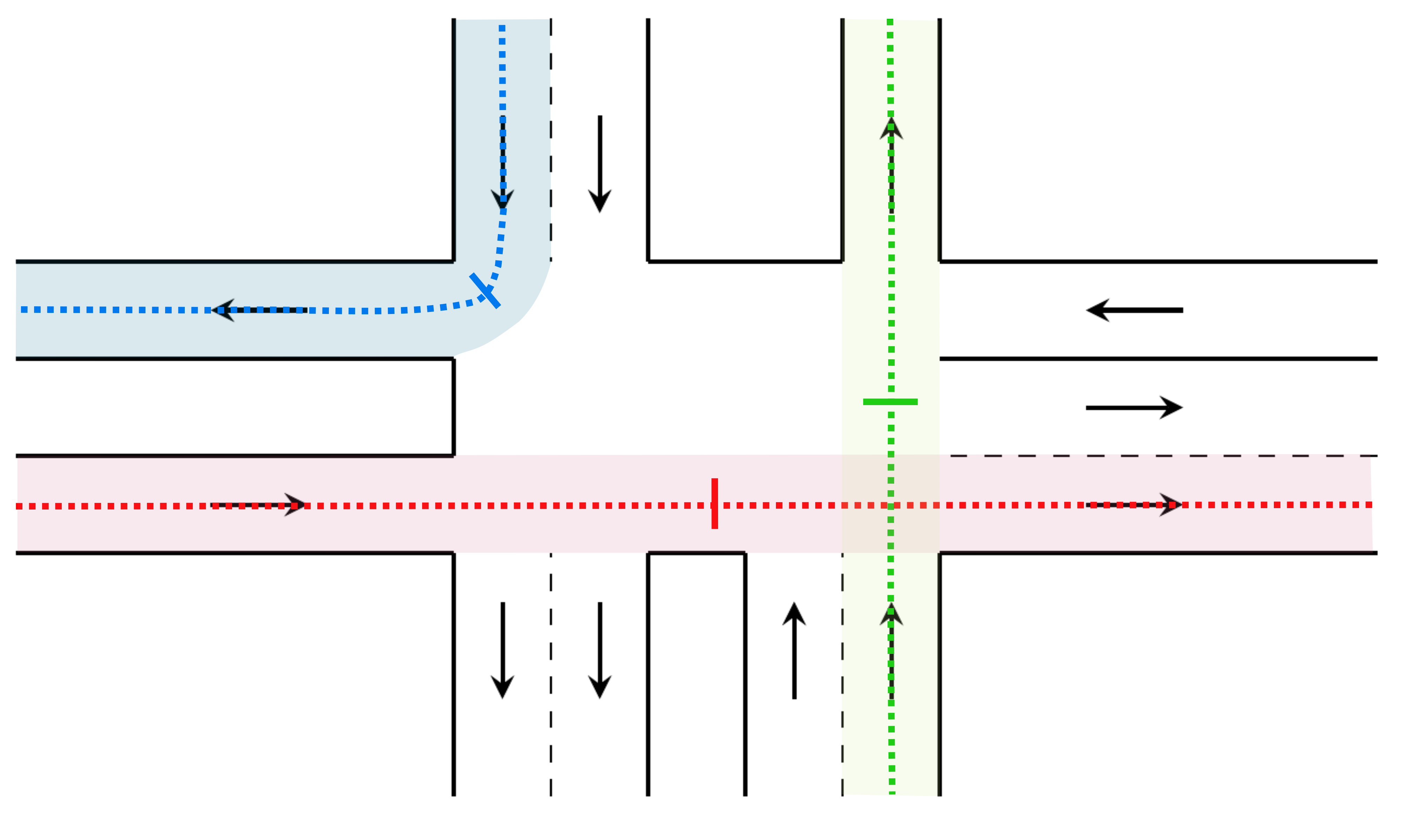}
		\caption{}
		\label{fig:4laneinitareas}
	\end{subfigure}	
\end{center}
	\caption{Initialization of the lanes as lanelets using the topology estimation. (a) Input trajectories. Colors highlight different track ids. (b) Split trajectories into two parts. (c) Assign each trajectory part to a lane. (d) Connect lanes assigned to the same trajectory with a straight interpolation.
	}
	\label{fig:lane_init}
\end{figure}

In the second step of this work, the input data needs to be enhanced with an association of object detections of the same vehicle. This means that every point detection gets associated with points from other time steps, yielding trajectories of objects. Using this information, we can connect estimated lanes at two different arms, in order to reconstruct a path that connects a pair of lanes on the intersection. As can be seen in \autoref{fig:steps_diagram}, different environment perceiving sensor types can be used for estimating the necessary trajectories when fed to a tracking algorithm. 

This step refines the coarse model estimated in the topology estimation (see Section \ref{sec:topo}) by estimating the lane course of the intersection, but it does not change the topology estimated before. 

\subsection{Preprocessing}
\label{sec:lane_preproc}
First, the trajectory data has to be processed as depicted in \autoref{fig:lane_init}. We split each trajectory into its parts as described in Section \ref{sec:topo_preproc} and depicted in \autoref{fig:2laneinittrajsplit}. Then, our algorithm assigns each trajectory part to the closest arm of the estimated intersection model by searching for the lane whose center line is the closest to the mean point of the trajectory part (see \autoref{fig:3laneinitsplitareas}). Each pair of trajectory parts can then be used to get possible connections between lanes (see \autoref{fig:4laneinitareas}). 

Each lane course is represented as a lanelet \cite{Poggenhans_Lanelet2highdefinitionmap_2018}, which is comprised of a pair of two poly lines, the left and the right border. Each lane in the topology model, estimated in the first step, is converted into a lanelet representation with equidistantly distributed support points (see \autoref{fig:teaser}). For every lane that is connected with another arm by a trajectory, we also connect the lanelets in order to have a continuous lane model within the intersection as depicted in \autoref{fig:4laneinitareas}. In the preprocessing step, the connection is a straight lanelet, but will be refined in the estimation. Adjacent lanes can share all or a subset of the support points of the lanelet border, depending on whether they are parallel over the whole intersection or either of them turns into another arm. 

For the generated lanelets we further calculate a center line for easier calculations later on. The center line $M_l$ of each lanelet is defined as the poly line of center points $m$ between opposing points on the lanelet borders $b \in B_l$. We also double the discretization rate of the center line by adding an additional support point between the existing ones.

We finally refine the representation using the trajectories. Each point of the center line is moved to minimize the distance to the assigned trajectories. As metric we calculate the orthogonal distance of the trajectory to the center line. 

With this refinement we finally calculate the neighbors of each border point that are candidates for fusing lane borders during the sampling. 

\subsection{Intersection Model}
The tracked object detections $T$ are a set of timestamped positions $p = (x, y)$ in a world-fixed coordinate system and an association that assigns multiple positions to a single trajectory $t_i$. Thus, we can additionally calculate the orientation $\vec{o}$ for each position. 

\begin{figure}[tb]
	\centering
	\includegraphics[width=1\linewidth]{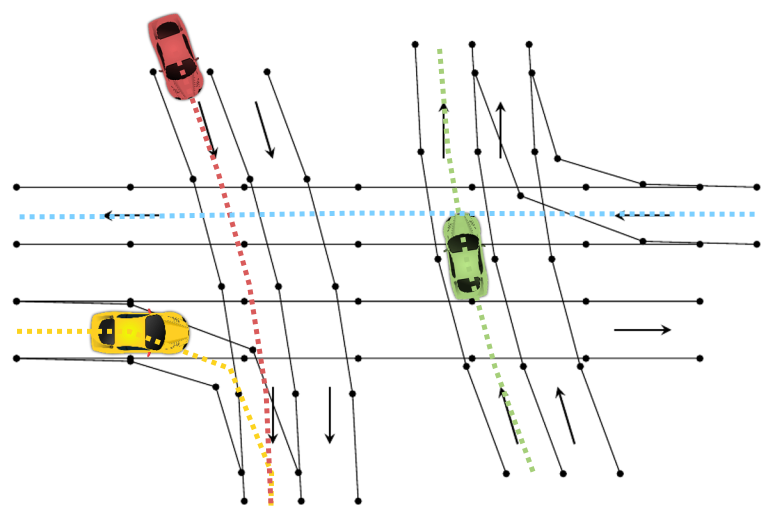}
	\caption{For the lane course estimation, the intersection is represented as a set of lanelets with equally distributed support points.}
	\label{fig:teaser}
\end{figure}

In this step, we model the intersection $I_2$ as set of lanelets $L$ as depicted in \autoref{fig:teaser}. Each lanelet $l$ consists of optionally shared border points $b \in B_l$ and center points $m \in M_l$. With this representation, each sampling step either 

\begin{itemize}
	\item modifies a point $m_l$ of the center line of a lane, 
	\item splits two adjacent lanelets by generating a new border point for one of them, which is sampled around the original position with distance $\Delta b \sim \mathcal{U}[\SI{0}{\meter},\SI{0.6}{\meter}]$ or
	\item merges two neighboring lanelets $l_a$ and $l_b$ by assigning both lanelets the same border point. The new border point is equally likely to be the one taken from $l_a$ or $l_b$. 
\end{itemize}

\subsection{Probabilistic Evaluation}
For evaluating the resulting lane courses of $I_2$ we calculate the posterior probability $P(I_2|T)$ depending on the detected trajectories $T$. As described in Section \ref{sec:topo}, this is split into calculating the prior on our model $P(I_2)$ and the likelihood for each trajectory point as $P(t_i|I_2)$. 

\subsubsection{Lane Course Prior}
The prior is based on the number of shared border points between two adjacent lanes and the smoothness of the lanes $s(I_2)^\tau$. The smoothness term is applied because intersections are machine or man made and both have a tendency to stay continuous when designing intersections. Vehicles are bound to non-holonomic movements that prevent roads from having huge curvature changes. 
We calculate the smoothness of each lane individually as sum of absolute angular errors. For each pair of three subsequent center points $\{i,j,k\} \in M_l$ we calculate the two direction vectors $\gamma_{ij}$ and $\gamma_{jk}$ and derive the sum of absolute differences in the angle between the directions as 
\begin{equation}
\delta_l = \sum_{i,j,k}^{M_l} |\arccos( \langle\gamma_{ij}, \gamma_{jk}\rangle )|.
\end{equation}
We assume this difference $\delta_l$ to follow a normal distribution. Using the two metrics, the prior is calculated as 
\begin{equation}
P(I_2)=s(I_2)^\tau \cdot \prod_{l \in L} P(\delta_l).
\end{equation}


\subsubsection{Likelihood}
For the fitness of the estimated lane course and detected trajectories, the distance between the trajectory points and the center line is calculated. We reuse the orthogonal distance $d_\perp$ from Section \ref{sec:lane_preproc} between a point on the center line $m \in M_l$ of lane $l$ and a trajectory point $p \in t_i$ and assume it to follow a normal distribution 
which leads to 
\begin{equation}
P(t_i|I_2) = \sum_{l\in L} \mathbf{1}(t_i,l) \prod_{p \in t_i}  P_\perp(p|M_l,I).
\end{equation}

%% file: content/experiments.tex
\section{Experimental Evaluation}
Because of the lack of publicly available datasets containing precise, lane-level maps, we created ground truth for 14 real-world intersections manually labeled in a map format. We separated those into small and big intersections. 
For a more meaningful, quantitative evaluation, we generated 1000 artificial intersections randomly using the parameters from \autoref{tab:sim_params}.

\begin{table}[tb]
	\caption{Parameters and their definition range for generating intersections.}
	\label{tab:sim_params}
	\begin{center}
		\begin{tabular}{l|l}
			\textbf{Parameter} & \textbf{Definition Range} \\
			\hline \\
			Number of arms 			 & $|A| \in [3,4,5]$ \\[2pt]
			Lanes per driving direction & $|L^i|, |L^o| \in [1,2,3,4]$ \\[2pt]
			Angle between arms 		 & $\alpha_{ab} \geq 45^\circ$ \\[2pt]
			Width of the gap 			 & $g_a < \SI{3}{\meter}$ \\[2pt]
		\end{tabular}
	\end{center}
\end{table}
For both simulated and real intersections, we simulated vehicles traveling alongside the lanes, which resulted in a maximum of six trajectories per lane. The vehicle routes have been determined by randomly choosing two lanes in the intersection and following the ground truth center lines. For a more realistic simulation, we added Gaussian noise with $\Delta d \leftarrow \mathcal{N}(\SI{0}{\meter},\SI{1}{\meter})$ to every detection and some random, false detections around the center with $s \leftarrow \mathcal{U}(\SI{0}{\meter},\SI{80}{\meter})$, in order to replicate sensor measurements, which are often prone to noise.

All following experiments were conducted on a Ubuntu system with an Intel Core i7-8750H CPU \SI{2.20}{\giga\hertz} (Turbo Boost: \SI{4.1}{\giga\hertz}). The implementation has been done in the ROS framework in C++, allowing for fast execution time. In order to make this approach feasible for online usage, we limit the evaluation to execution times of approximately \SI{150}{\milli\second}, which results in limiting the number of samples that each step is allowed to generate, before the result for evaluation is drawn.

\subsection{Topology Estimation}
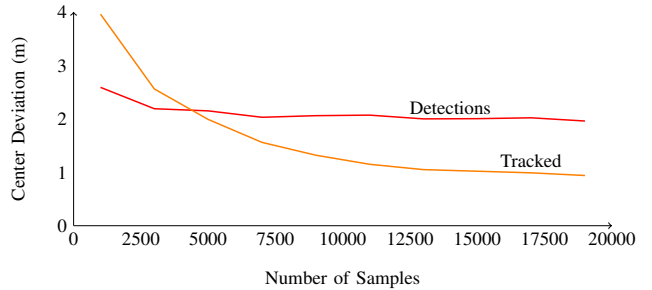
\begin{figure}[tb]
	\centering
	\resizebox{\linewidth}{!}{
		\input{figures/num_samples_acc_first_step.tex}
	}
	\caption{Topology Estimation: Influence of number of samples on the accuracy of the intersection center.}
	\label{fig:num_samples_vs_center}
\end{figure}

First, we evaluate the topology estimation.
We use distance measures on the individual parameters (see Section \ref{sec:topo_model}) to determine the quality of the estimated intersection.

We found that our approach is able to detect the topology, describing the estimated number of arms, for all synthetic intersections correctly although we added noise and simulated false detections. Only on a single intersection (in the raw detections experiment), we could not detect the correct number of arms, which led to an accuracy of \SI{99.90}{\percent}.

The more complex lane-level topology, where the number of lanes and their direction in each arm is regarded but not the precise course, achieved an accuracy of \SI{92.28}{\percent} when executed on tracked data. Based on raw detections we could estimate the lane-level topology with an accuracy of \SI{84.90}{\percent}. The errors on the lane-level occurred because an additional lane at the outer borders of the models was estimated in some cases.

\begin{figure}[tb]
	\centering
	\resizebox{\linewidth}{!}{
		\input{figures/num_samples_exec_acc_snd_step.tex}
	}
	\caption{Influence of the number of samples on the execution time.}
	\label{fig:num_samples_vs_exec}
\end{figure}
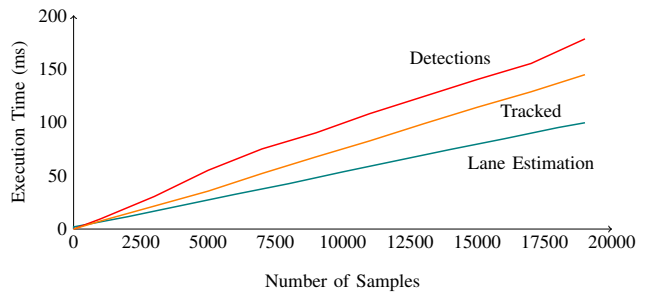

With 5000 samples we achieved an average execution time of \SI{55.38}{\milli\second} in case of raw detections and \SI{35.72}{\milli\second} in case of tracked objects. The difference can be explained by the considerable reduction of points in the tracked case. When changing the number of samples for an estimation the accuracy of e.g. the center (see \autoref{fig:num_samples_vs_center}) improves exponentially, whereas the execution time changes linearly (see \autoref{fig:num_samples_vs_exec}). 

\begin{figure}[tb]
	\begin{center}
		\begin{subfigure}{0.492\linewidth}
			\centering
			\includegraphics[width=1\linewidth]{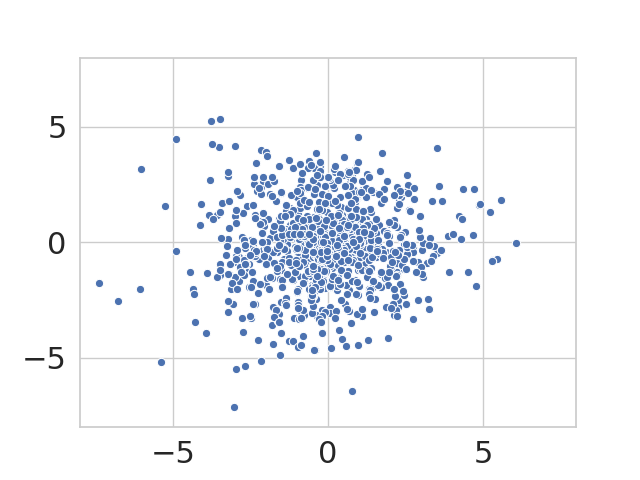}
			\caption{Data: detections}
			\label{fig:centerscatter_radar}
		\end{subfigure}
		\begin{subfigure}{0.492\linewidth}
			\centering
			\includegraphics[width=1\linewidth]{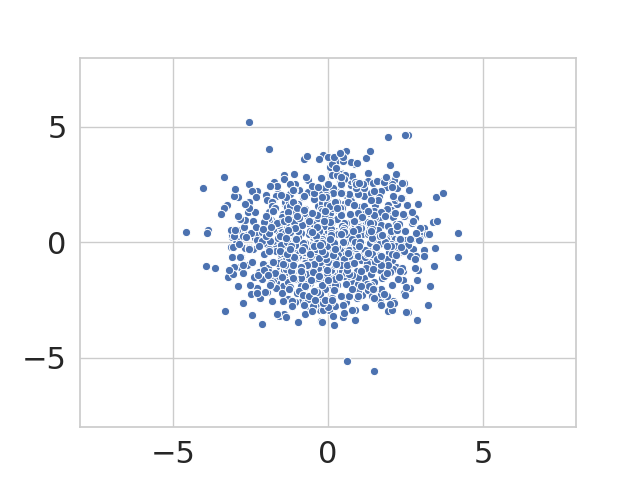}
			\caption{Data: tracked objects}
			\label{fig:centerscatter_traj}
		\end{subfigure}
	\end{center}
	\caption{Topology Estimation: Distribution of the error on the center position after \SI{5000}{} samples. For the raw detection case, three outliers of $(1,-20)$, $(9,-19)$ and $(1,21)$ are outside of the plot.}
	
	\label{fig:centerscatter}
\end{figure}

For the center of the intersections we depicted the error distribution in \autoref{fig:centerscatter}. Here, we can infer that the distribution is a consequence of the noise, that we added to our measurements.
Additionally the center is prone to be wrongly estimated, without leading to wrong intersections in terms of driveability. Since moving the center of the intersection influences the direction of the arms, a wrong center position might compensate errors in the lane angle estimation.

For most cases of tracked data, the road angles are less than \ang{1} with an average error of \ang{0.34}. In the detection case, we estimate the angle with a deviation of \ang{0.97} on average.

In summary, our results show, that the topology estimation benefits from fewer detections and a tracking in beforehand, which increases the accuracy and speed of the approach. 

\subsection{Lane Course Estimation}
For evaluating the course of the individual lanes, we used a different measure.
We calculated the orthogonal distance between the estimated center line and the ground truth center line for all correctly estimated lanes. The overall quality of a lane was formulated as the mean Euclidean distance. The execution times presented here also include the preprocessing described in Section \ref{sec:lane_preproc}.

For the synthetic intersections, we achieved an accuracy of \SI{14}{\centi\meter} on average.
\SI{20000}{} sample points allowed for an execution time of \SI{104.05}{\milli\second}.  
With \SI{5000}{} samples in the topology estimation and \SI{20000}{} for the lane course, we achieve an overall execution time of \SI{140.05}{\milli\second}. When choosing fewer samples the execution time is reduced as depicted in \autoref{fig:num_samples_vs_exec}, with exponentially worse results.

\begin{figure*}
	\begin{center}
		\begin{subfigure}{0.3\linewidth}
			\centering
			\includegraphics[width=1\linewidth]{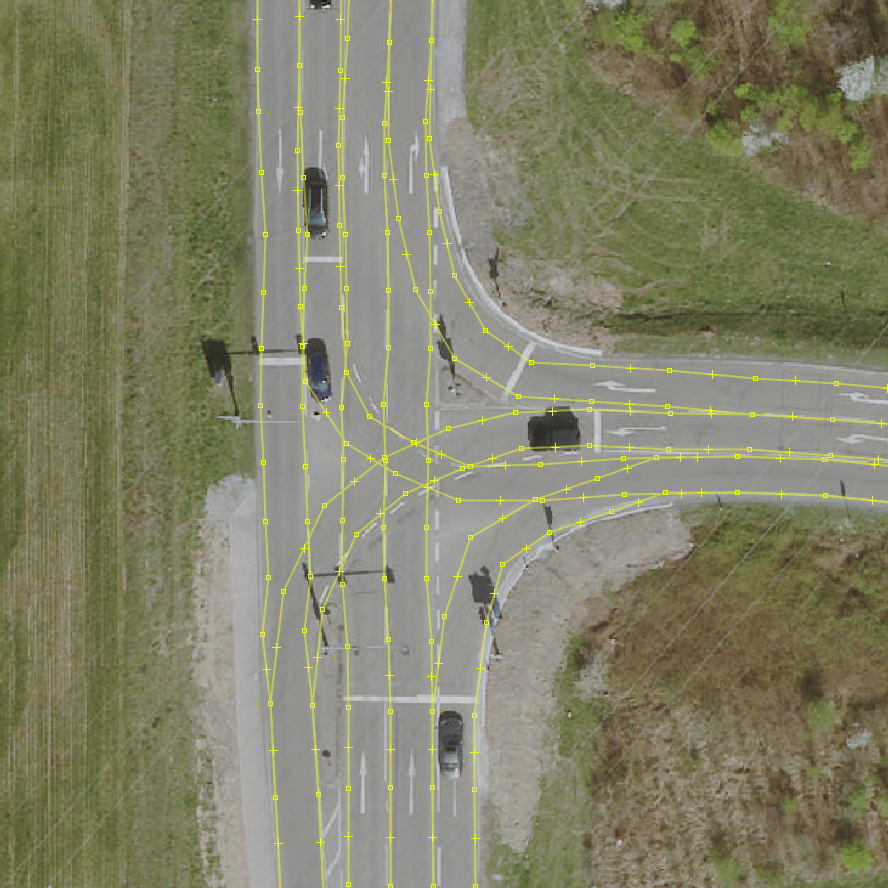}
		\end{subfigure}
		\begin{subfigure}{0.3\linewidth}
			\centering
			\includegraphics[width=1\linewidth]{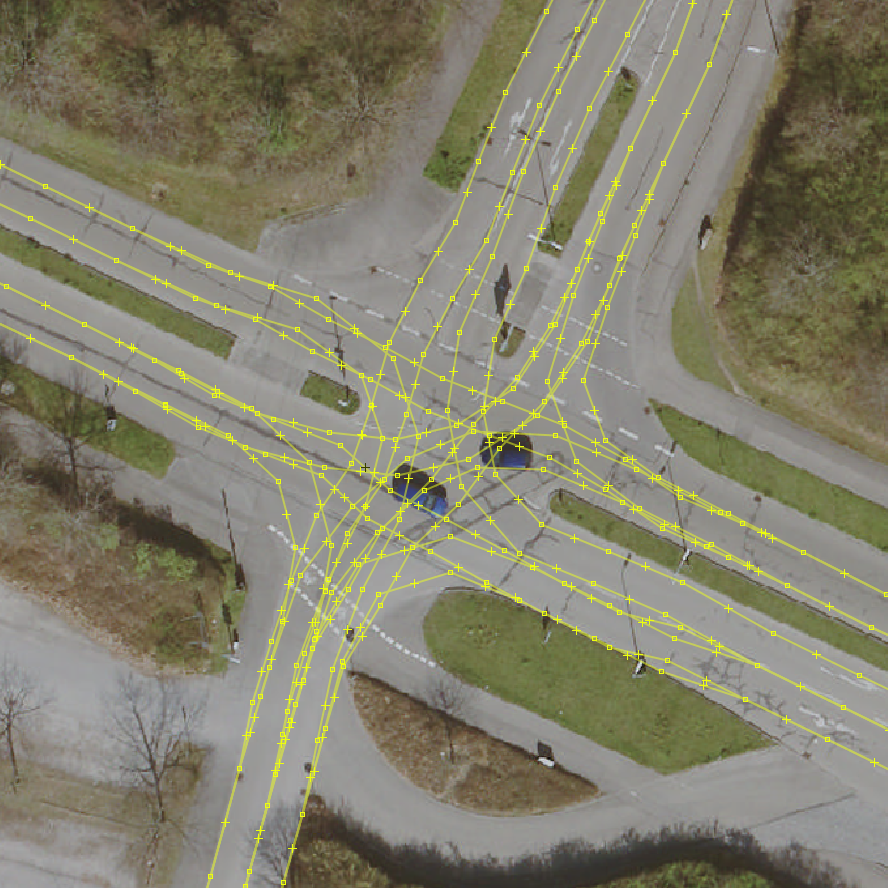}
		\end{subfigure}
		\begin{subfigure}{0.3\linewidth}
			\centering
			\includegraphics[width=1\linewidth]{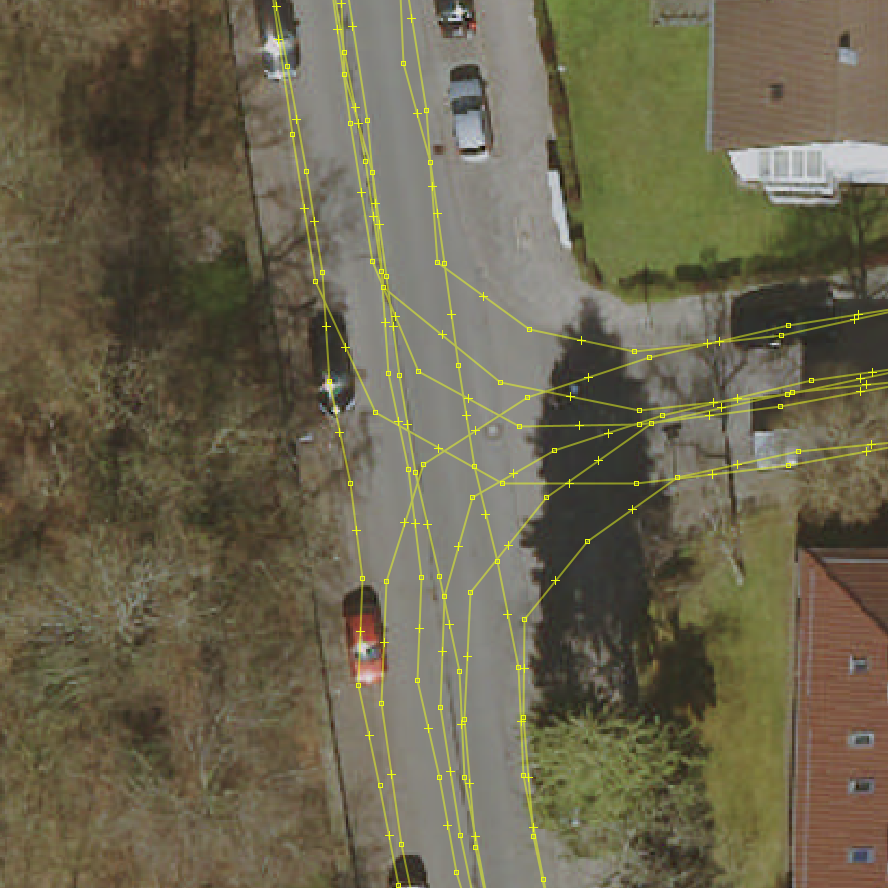}
		\end{subfigure}
		\begin{subfigure}{0.3\linewidth}
			\centering
			\includegraphics[width=1\linewidth]{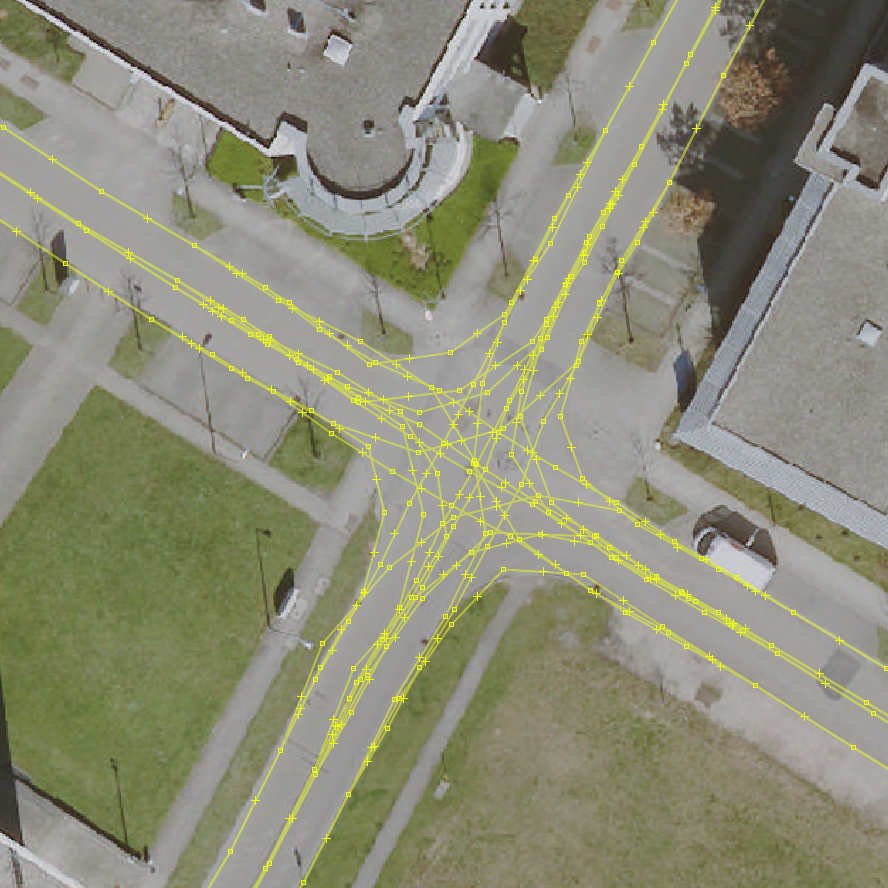}
		\end{subfigure}
		\begin{subfigure}{0.3\linewidth}
			\centering
			\includegraphics[width=1\linewidth]{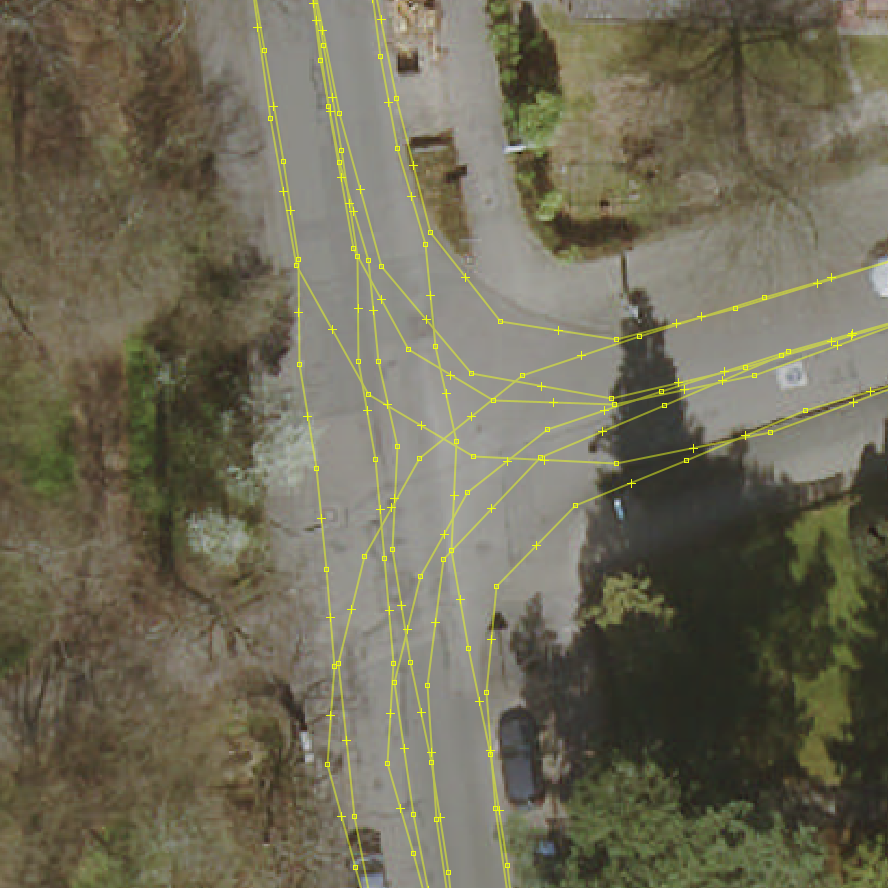}
		\end{subfigure}
		\begin{subfigure}{0.3\linewidth}
			\centering
			\includegraphics[width=1\linewidth]{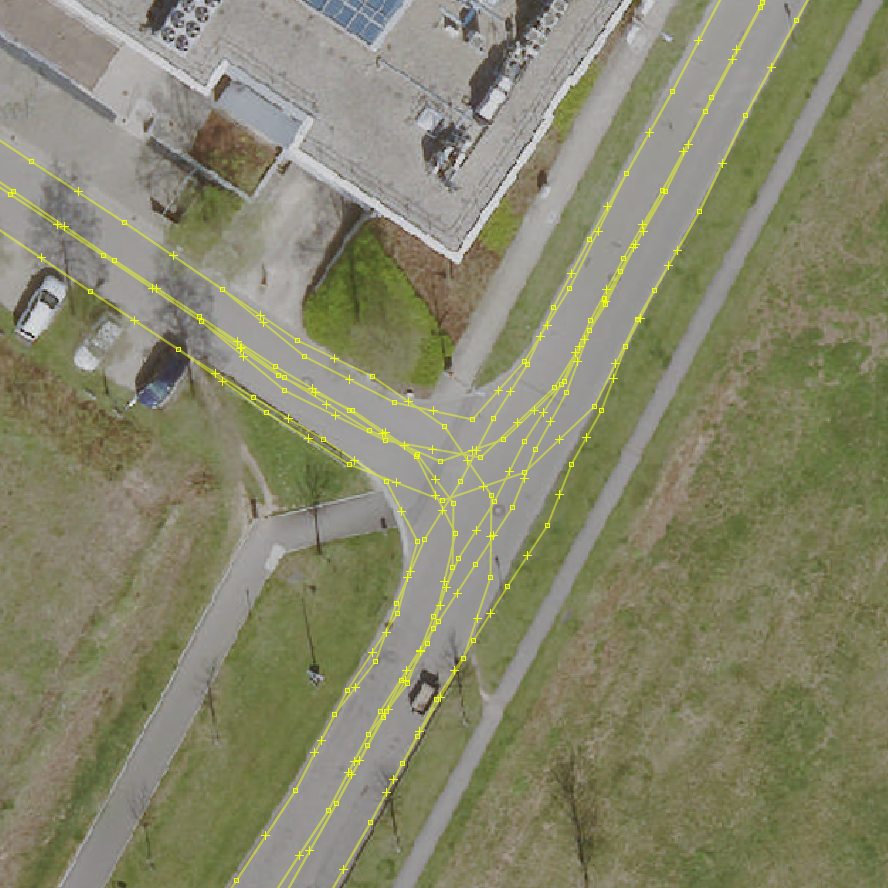}
		\end{subfigure}
		\begin{subfigure}{0.3\linewidth}
			\centering
			\includegraphics[width=1\linewidth]{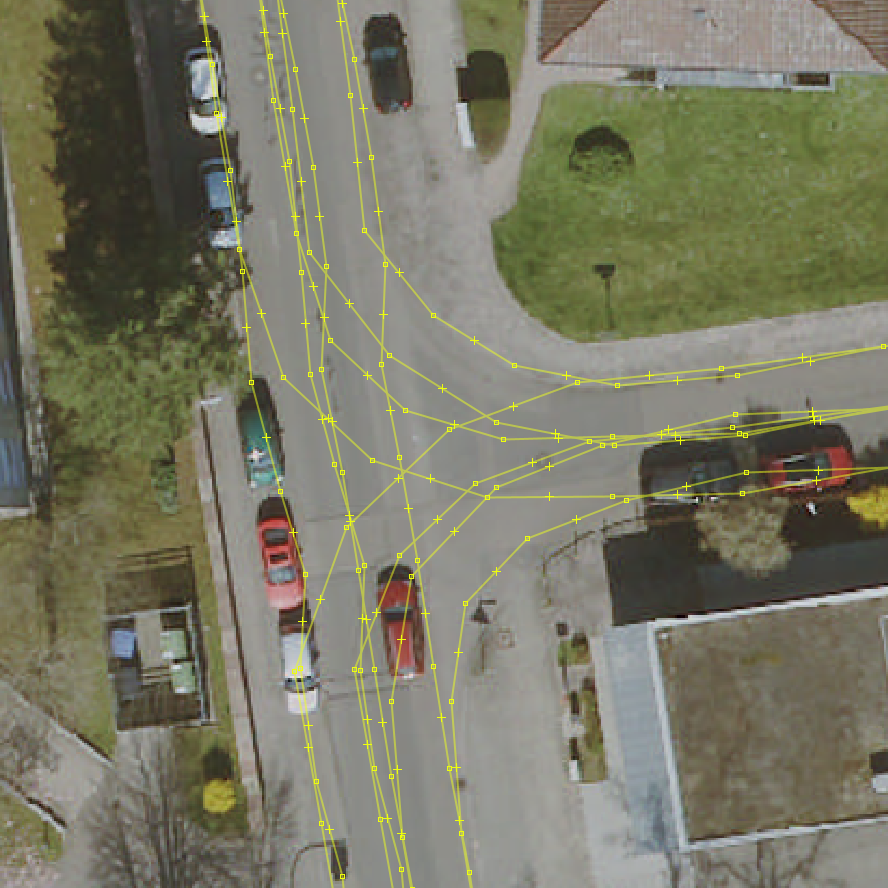}
		\end{subfigure}
		\begin{subfigure}{0.3\linewidth}
			\centering
			\includegraphics[width=1\linewidth]{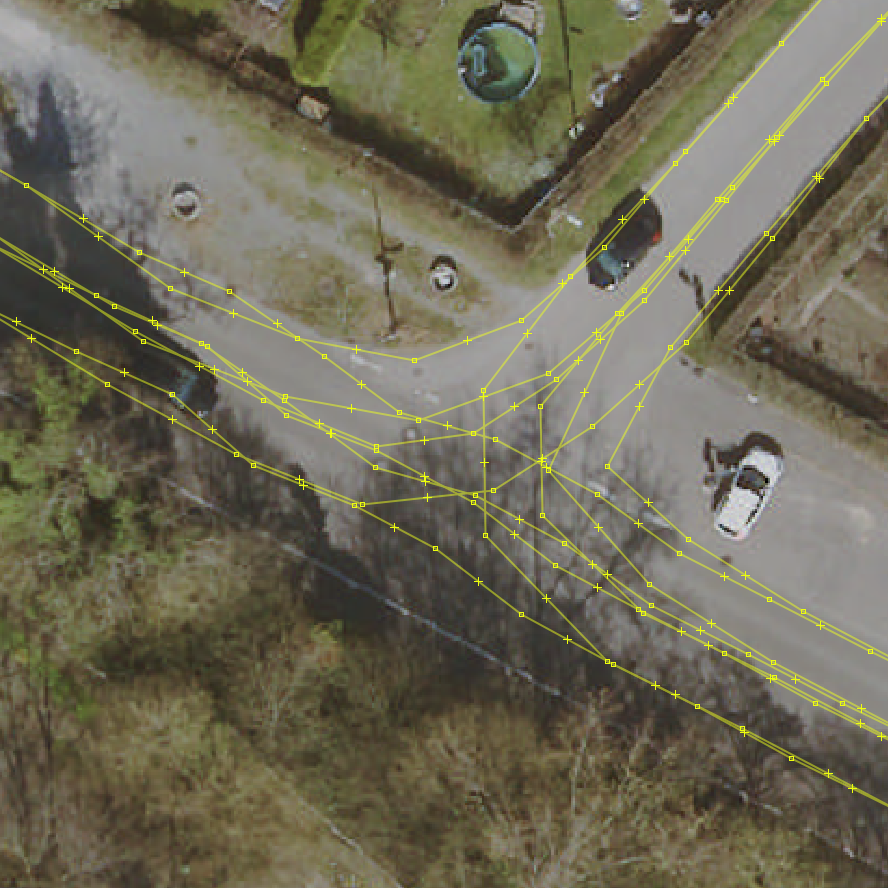}
		\end{subfigure}
		\begin{subfigure}{0.3\linewidth}
			\centering
			\includegraphics[width=1\linewidth]{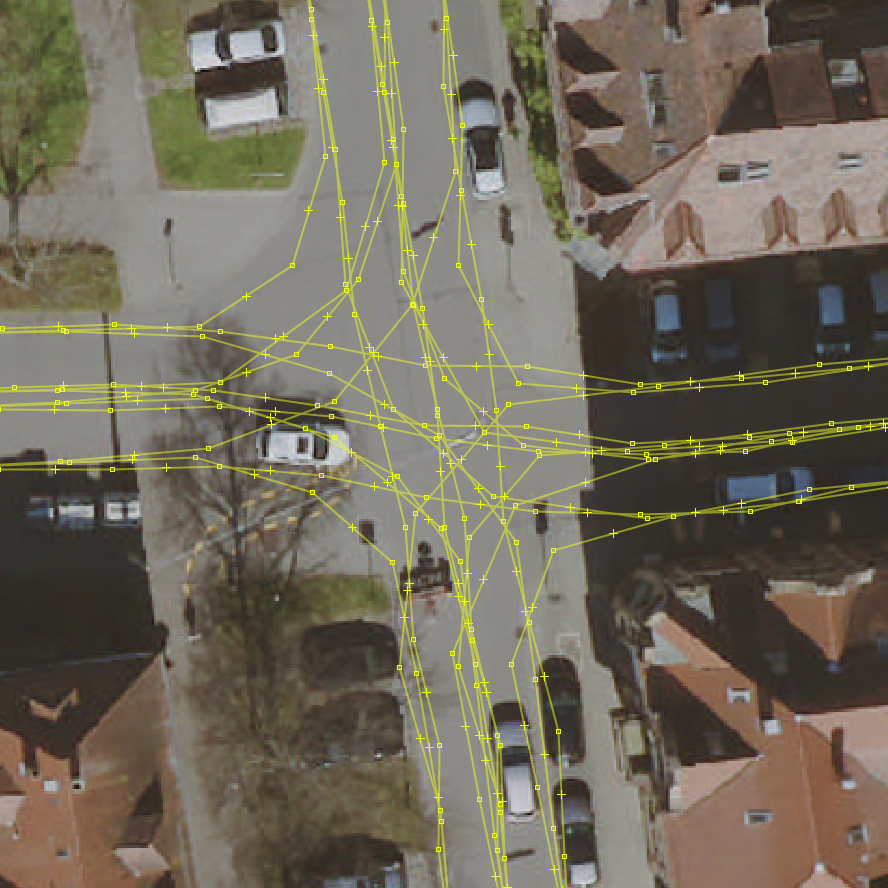}
		\end{subfigure}
	\end{center}
	\caption{Example results of real-world intersections. Aerial images: City of Karlsruhe, www.karlsruhe.de, dl-de/by-2-0}
	\label{fig:osm}
\end{figure*}

In the real-world scenarios we also used \SI{20000}{} sample points and achieved an accuracy of \SI{18}{\centi\meter} for the small intersections and \SI{27}{\centi\meter} for big ones. For smaller intersections, we get the results after \SI{77.5}{\milli\second} on average, whereas for big ones after \SI{89}{\milli\second}. The real scenarios could be solved quite faster, since we usually have fewer lanes per arm, than were simulated.
For a qualitative evaluation of the real scenarios, we depicted examples in \autoref{fig:osm}.

%% file: figures/num_samples_acc_first_step.tex
\begin{tikzpicture}


\draw[->] (0,0) -- (10,0) node[anchor=south] {};
\draw (5,-1) node {Number of Samples}; 
\draw	(0,0) node[anchor=north] {0}
(1.25,0) node[anchor=north] {2500}
(2.5,0) node[anchor=north] {5000}
(3.75,0) node[anchor=north] {7500}
(5,0) node[anchor=north] {10000}
(6.25,0) node[anchor=north] {12500}
(7.5,0) node[anchor=north] {15000}
(8.75,0) node[anchor=north] {17500}
(10,0) node[anchor=north] {20000};

\draw[->] (0,0) -- (0,4) node {};
\draw (-1,2) node[rotate=90] {Center Deviation (m)}; 
\draw (0,0) node[anchor=east] {0}
(0,1) node[anchor=east] {1}
(0,2) node[anchor=east] {2}
(0,3) node[anchor=east] {3}
(0,4) node[anchor=east] {4};


\draw[thick, red] (0.5, 2.59) -- (1.5, 2.19)-- (2.5, 2.15)-- (3.5, 2.03)-- (4.5, 2.06) -- (5.5, 2.07)-- (6.5, 2.00)-- (7.5, 2.004)-- (8.5, 2.02)-- (9.5, 1.96);
\draw (7,2.2) node {Detections}; 

\draw[thick, orange] (0.5, 3.96) -- (1.5, 2.56)-- (2.5, 1.99)-- (3.5, 1.56)-- (4.5, 1.32) -- (5.5,1.15)-- (6.5, 1.05)-- (7.5, 1.02)-- (8.5, 0.99)-- (9.5, 0.94);
\draw (8.5,1.21) node {Tracked}; 

\end{tikzpicture}

%% file: figures/num_samples_exec_acc_snd_step.tex
\begin{tikzpicture}


\draw[->] (0,0) -- (10,0) node[anchor=south] {};
\draw (5,-1) node {Number of Samples}; 
\draw	(0,0) node[anchor=north] {0}
(1.25,0) node[anchor=north] {2500}
(2.5,0) node[anchor=north] {5000}
(3.75,0) node[anchor=north] {7500}
(5,0) node[anchor=north] {10000}
(6.25,0) node[anchor=north] {12500}
(7.5,0) node[anchor=north] {15000}
(8.75,0) node[anchor=north] {17500}
(10,0) node[anchor=north] {20000};

\draw[->] (0,0) -- (0,4) node {};
\draw (-1,2) node[rotate=90] {Execution Time (ms)}; 
\draw (0,0) node[anchor=east] {0}
(0,1) node[anchor=east] {50}
(0,2) node[anchor=east] {100}
(0,3) node[anchor=east] {150}
(0,4) node[anchor=east] {200};


\draw[thick, teal] (0,0) -- (0.0005, 0.04) -- (1, 0.23) -- (2, 0.44) -- (3, 0.65) -- (4, 0.85) -- (5, 1.07) -- (6, 1.28) -- (7, 1.49) -- (8, 1.69) -- (9, 1.90) -- (9.5, 1.99);
\draw (8.5,1.25) node {Lane Estimation}; 

\draw[thick, red] (0,0) -- (0.5, 0.19) -- (1.5, 0.61)-- (2.5, 1.1)-- (3.5, 1.5)-- (4.5, 1.8) -- (5.5, 2.16)-- (6.5, 2.48)-- (7.5, 2.8)-- (8.5, 3.1)-- (9.5, 3.56);
\draw (7,3.2) node {Detections}; 

\draw[thick, orange] (0,0) -- (0.5, 0.15) -- (1.5, 0.43)-- (2.5, 0.71)-- (3.5, 1.04)-- (4.5, 1.35) -- (5.5,1.65)-- (6.5, 1.97)-- (7.5, 2.28)-- (8.5, 2.57)-- (9.5, 2.89);
\draw (8.5,2.21) node {Tracked}; 

\end{tikzpicture}

%% file: content/conlcusion.tex
\section{Conclusions}
In this work, we showed an approach for estimating both the coarse lane-level intersection topology and the lane course inside and outside the intersection. 
On simulated data, the topology was estimated correctly in all cases based on trajectories of other traffic participants. For simulated and real-world intersections the lanes borders were detected with an average error of \SI{14}{\centi\meter} and \SI{23}{\centi\meter}, resp. We are able to calculate the results within \SI{140}{\milli\second} and \SI{113}{\milli\second}. 
However, we chose this approach, because we can extract results at any time during the estimation, risking, of course, a less precise result. 

Our model is able to represent a lot of different intersections because it is neither limited in the number nor the geometry of arms and their lanes. To our knowledge, similar approaches either could not achieve comparable results or require considerably more computation time.

Our method of estimating intersection topologies on the fly overcomes critical limitations of state-of-the-art autonomous diving solutions. It allows to check the validity of maps, to update outdated maps and even to drive in unknown environments.

Since our approach is based on trajectory data, it can also be used in dense traffic when traditional features for roadway recognition, e.g. markings or curbstones, fail due to occlusions. However, we plan to extend our framework to those traditional features in order to benefit from multiple independent features which promises to increase robustness even further. 